\documentclass[runningheads]{llncs}
\usepackage[T1]{fontenc}
\usepackage{graphicx}
\usepackage{amsmath}
\usepackage{amssymb}
\usepackage{booktabs}
\usepackage{multirow}

\newcommand{\ie}{{\em i.e.}}
\newcommand{\etal}{{\em et~al.}}
\newcommand{\eg}{{\em e.g.}}

\begin{document}
\title{Fragile by Design: On the Limits of Adversarial Defenses in Personalized Generation}
\titlerunning{AntiDB\_Purify}
%
\author{Zhen Chen\inst{1}\and
Yi Zhang\inst{2}\and
Xiangyu Yin\inst{1}\and
Chengxuan Qin\inst{1}\and
Xingyu Zhao\inst{2}\and
Xiaowei Huang\inst{1}\and
Wenjie Ruan\inst{1}
}
\authorrunning{Z. Chen et al.}
%
\institute{Department of Computer Science, University of Liverpool, Liverpool, L69 3BX, UK \and
WMG, University of Warwick, Coventry, CV4 7AL, UK\\
}

\maketitle              
\begin{abstract}
Personalized AI applications such as DreamBooth enable the generation of customized content from user images, but also raise significant privacy concerns, particularly the risk of facial identity leakage. Recent defense mechanisms like Anti-DreamBooth attempt to mitigate this risk by injecting adversarial perturbations into user photos to prevent successful personalization. However, we identify two critical yet overlooked limitations of these methods. First, the adversarial examples often exhibit perceptible artifacts such as conspicuous patterns or stripes, making them easily detectable as manipulated content. Second, the perturbations are highly fragile, as even a simple, non-learned filter can effectively remove them, thereby restoring the model's ability to memorize and reproduce user identity. To investigate this vulnerability, we propose a novel evaluation framework, \textbf{AntiDB\_Purify}, to systematically evaluate existing defenses under realistic purification threats, including both traditional image filters and adversarial purification. Results reveal that none of the current methods maintains their protective effectiveness under such threats. These findings highlight that current defenses offer a false sense of security and underscore the urgent need for more imperceptible and robust protections to safeguard user identity in personalized generation. Code is available at \url{https://github.com/TrustAI/AntiDB\_Purify}.

\end{abstract}

\section{Introduction}
The era of rapidly advancing AI generative models, especially the emergence of diffusion models (DMs)~~\cite{song2020denoising,dhariwal2021diffusion,rombach2022high}, has significantly improved the realism and diversity of synthesized images~\cite{dhariwal2021diffusion}. The text-to-image generation techniques offer greater convenience for image generation, particularly when combined with large-scale, pre-trained models. Recently, personalized AI~\cite{galimage,ruiz2023dreambooth,kumari2023multi} has become feasible, enabling users to generate art content for a specific subject or object. One prominent technique for this purpose is DreamBooth~\cite{ruiz2023dreambooth}, which has enabled powerful text-to-image synthesis using only a few personal portrait images by fine-tuning Stable Diffusion~\cite{stablediffusion2022}. It is widely used in applications such as virtual avatars, fan art, and customized media generation~\cite{zhou2024customization,jin2024customized}. However, this convenience introduces significant privacy concerns~\cite{sundar2010personalization,arlein2000privacy}. Once a user's face is used to fine-tune such a model, its identity can be memorized and reproduced indefinitely, often without their consent.

To address this issue, several recent works have proposed anti-personalization defenses~\cite{van2023anti,onikubo2024high,wang2024simac,liu2024disrupting,song2025idprotector,wan2024prompt}. These methods aim to protect users by injecting adversarial perturbations into their clean photos, making it difficult for DreamBooth to reproduce the user’s identity. Ideally, these perturbations should be imperceptible to humans while still effective in preventing identity learning. While achieving impressive performance under controlled conditions, we identify two critical and underexplored weaknesses shared by most existing defenses:

\begin{figure*}[t]
\centering
\includegraphics[width=1\textwidth]{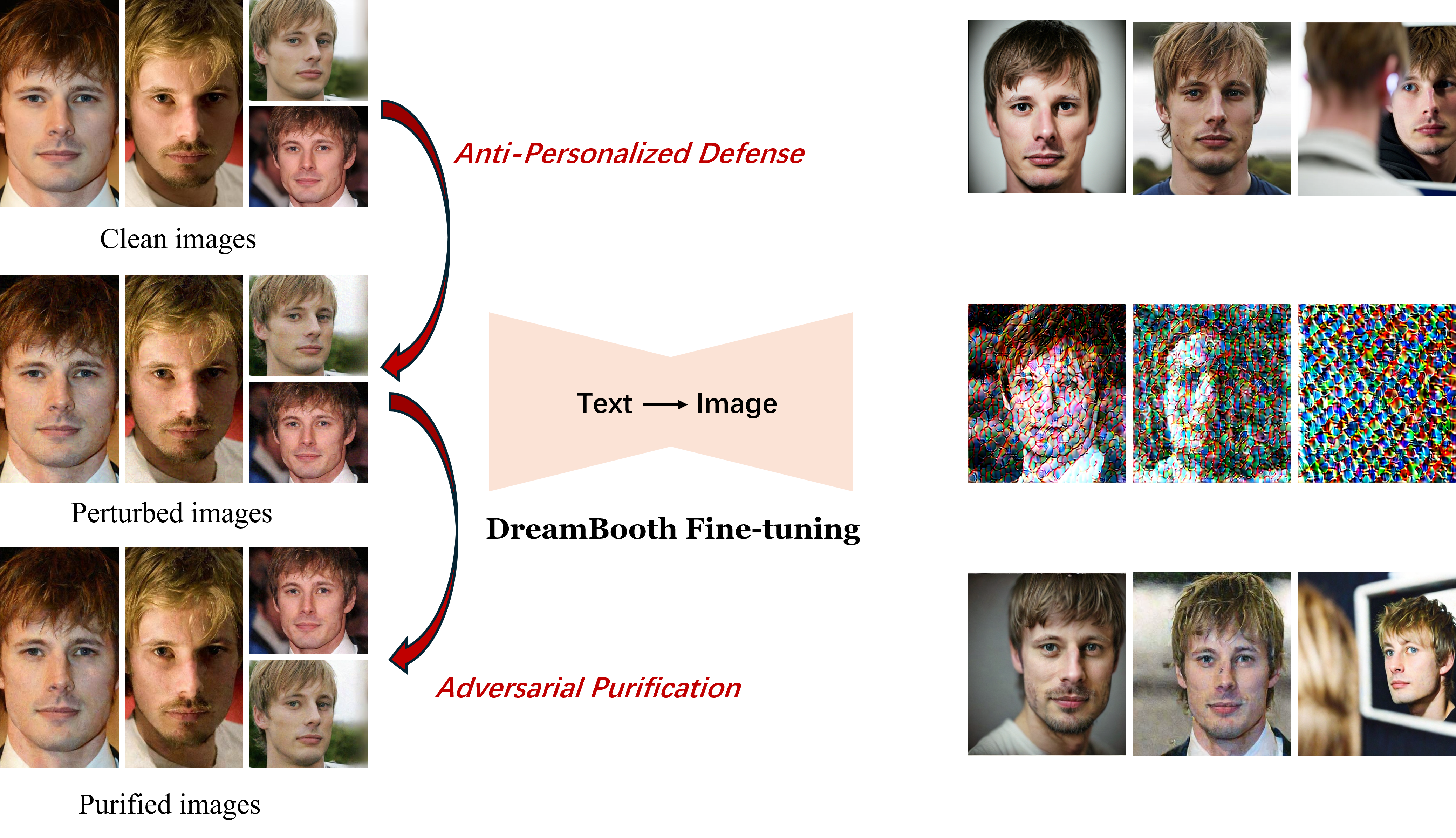} 
\caption{Purification can remove adversarial perturbations, thereby undermining the effectiveness of anti-personalization.}
\label{intro}
\end{figure*}

\textbf{Perceptibility}: Most adversarial examples exhibit perceptible artifacts, such as stripe-like noise, which compromise their imperceptibility and can be easily spotted as manipulated or fake images. Thus, they are susceptible to detection and removal by potential adversaries.

\textbf{Filtering Fragility}: Basic image processing operations, such as Gaussian blur, bilateral filtering~\cite{tomasi1998bilateral}, or diffusion-based denoising~\cite{nie2022diffusion,yoon2021adversarial,zhao2024can} can nullify the adversarial noises they inject. Notably, some filters do not require machine learning expertise, making them highly accessible and easy to deploy in practical scenarios. Consequently, they are fragile under post-process purification.

These two properties expose a \textbf{false sense of security in existing defenses}. In practice, an adversary might unintentionally obtain user photos embedded with adversarial perturbations. The presence of such perturbations may be visually apparent due to noticeable artifacts or may be inferred from the failure of DreamBooth personalization. In either case, the attacker could apply a simple purification technique to obtain purified images, which can then be used to fine-tune a DreamBooth model to reconstruct the user's identity, completely bypassing defense mechanisms. As shown in Fig.~\ref {intro}, the adversarial examples generated by anti-personalization defenses can protect the user's identity. However, after applying adversarial purification, DreamBooth’s fine-tune is still able to learn and reproduce identity-specific features. 

To the best of our knowledge, only HF-ADB (High-Frequency Anti-DreamB\\-ooth)~\cite{onikubo2024high} attempts to mitigate this challenge by adding stronger perturbations to the high-frequency regions of an image. The intuition is that high-frequency signals are harder to remove. However, our results show that this strategy fails to provide effective protection even before purification, let alone after. Nevertheless, the perturbations of HF-ADB exhibit higher resistance to purification, which may be a promising direction toward addressing this challenge in future work.

In this paper, we present the first comprehensive evaluation framework, \textbf{AntiDB\_Purify}, to systematically examine anti-personalized defenses under realistic purification threats. Specifically, we evaluate four state-of-the-art defense methods against both traditional filtering techniques and adversarial purifications. Our findings reveal that none of the current defenses retain their protective effect after purification, and all fail to prevent identity-specific personalization by DreamBooth. These findings highlight an urgent and unresolved challenge in the field of privacy-preserving personalized generation. Our main contributions can be summarized as follows.

\begin{itemize}
    \item We identify two critical limitations: perceptibility and filtering fragility in existing anti-personalization defenses, which have been largely overlooked in prior work.

    \item We recommend a more realistic and rigorous evaluation framework for anti-personalization defenses, reflecting realistic threat scenarios in which adversaries perform purification before fine-tuning generative models.

    \item We empirically demonstrate that existing defenses fail to retain their effectiveness after purification, highlighting the urgent need for more robust protection mechanisms in personalized generative models.
\end{itemize}

\section{Related Works}
\subsection{Personalized Generative Models}
Text-to-image diffusion models have revolutionized generative AI by enabling the synthesis of high-quality and diverse images conditioned on textual descriptions~\cite{zhang2023adding,saharia2022photorealistic}. These models leverage iterative denoising processes to generate images that are semantically aligned with the input prompts. Stable Diffusion is a prominent approach that operates in a learned latent space rather than pixel space. The design is rooted in Latent Diffusion Models (LDMs)~\cite{rombach2022high}, significantly improving computational efficiency while maintaining high generation fidelity.

Consequently, personalized text-to-image generation has emerged as a powerful capability of diffusion models, enabling the synthesis of identity-consistent images from a few user images. Earlier approaches, such as Textual Inversion~\cite{galimage}, achieve personalization by optimizing pseudo-token embeddings that encode subject identity within the prompt space. DreamBooth~\cite{ruiz2023dreambooth} has emerged as the most widely adopted method for personalized image generation. It fine-tunes a pre-trained diffusion model on a small set of subject-specific images while associating the identity with a unique textual token. Due to its strong ability to preserve facial identity across a wide range of prompts with high visual fidelity, DreamBooth has been extensively integrated into user-facing applications and embraced by open-source communities.
More recently, methods such as LoRA-based~\cite{hulora} fine-tuning and Custom Diffusion~\cite{kumari2023multi} are inspired by DreamBooth, employing parameter-efficient techniques that modify internal model representations for enhanced subject fidelity and improve generalization across diverse prompts.

\subsection{Adversarial Defenses against DreamBooth}
Adversarial robustness~\cite{szegedy2013intriguing,goodfellow2014explaining} has long been a central topic in the study of AI safety and continues to receive attention in recent years~\cite{chen2024nrat,chen2024tarp,zhang2024prass,zhang2024towards,zhang2024protip,zhang2025adversarial,yin2024boosting,wang2025black}, encompassing adversarial attacks and defenses that reveal and mitigate model vulnerabilities. Inspired by adversarial attacks~\cite{dong2018boosting,huang2017adversarial,wang2025black} in classification tasks, Van \etal~\cite{van2023anti} proposed Anti-DreamBooth, which is the first defense method specifically designed to disrupt the personalization of DreamBooth. It leverages a surrogate DreamBooth model to iteratively optimize adversarial noise and finally generates adversarial examples that disrupt the personalization process of DreamBooth. HF-ADB~\cite{onikubo2024high} further decomposes an image into high-frequency and low-frequency regions, and applies stronger perturbations specifically to the high-frequency components. SimAC~\cite{wang2024simac} leverages a greedy approach to identify and select the most effective timesteps for perturbation, replacing the random timestep selection strategy used in Anti-DreamBooth. DisDiff~\cite{liu2024disrupting} disrupts the internal image-text alignment by targeting the textual guidance mechanism. In addition, it integrates diffusion sampling with adversarial optimization via a merit sampling scheduler, which adaptively constrains the perturbation update magnitude.

\subsection{Purification Techniques}
Traditional purification methods include simple filtering techniques such as bilateral filtering~\cite{tomasi1998bilateral}, which smooths high-frequency adversarial noise while preserving important image structures. Guided filtering~\cite{he2015fast} assumes that the output image can be locally modeled as a linear transformation of a guidance image, thereby enabling edge-preserving smoothing. When combined, these filters can effectively restore clean image features and further enhance the purification outcome. Their lightweight nature and ease of deployment make them practical tools for image purification in real-world scenarios.

Adversarial purification methods aim to eliminate adversarial perturbations by projecting inputs back onto the manifold of clean data, typically through denoising or reconstruction processes. In contrast to those require retraining or modification of the target model, \eg, adversarial training~\cite{shafahi2019adversarial,ganin2016domain}, purification techniques are generally employed as post-processing steps and can be applied independently of the downstream model architecture. This flexibility makes purification techniques attractive for defending against various adversarial attacks in practical settings. Recent advances in adversarial purification have primarily focused on diffusion-based methods. Among them, DiffPure~\cite{nie2022diffusion} is the prominent approach that leverages the reverse diffusion process of a pre-trained generative model to effectively remove adversarial noise. It demonstrates strong generalization across various datasets and attack types. GrIDPure~\cite{zhao2024can} builds upon it by incorporating a guided iterative denoising process, which integrates intermediate reconstructions with adversarial-aware sampling to enhance robustness. Both methods represent the state-of-the-art in purification techniques, exhibiting strong capabilities in mitigating carefully crafted adversarial perturbations.

\section{Purification as a Threat: Revisiting Anti-Personalization Defenses}

In this section, we first formally define our problem and review existing anti-personalization methods, highlighting their common design patterns. We then introduce purification techniques and analyze the vulnerabilities of current anti-personalization defenses under such purifications.

\subsection{Problem Definition}

Building upon the existing formulation of the Anti-DreamBooth problem, we introduce a more realistic constraint wherein an adversary applies purification to filter out adversarial perturbations before DreamBooth fine-tuning.

Let $\mathcal{X}$ denote the set of user images intended for protection, where each image $x \in \mathcal{X}$ contains identity-revealing features. An anti-personalization defense constructs adversarially perturbed images $x' = x + \delta$, with $\delta$ being a bounded adversarial noise (e.g., $\| \delta \|_p \leq \eta$), and publishes the perturbed set $\mathcal{X}' = \{x + \delta\}$, while keeping the original $\mathcal{X}$ private. A realistic attacker may collect a small subset $\mathcal{X}'_{\text{db}} = \{x^{(i)} + \delta^{(i)}\}_{i=1}^{N_{\text{db}}} \subset \mathcal{X}'$ to fine-tune a pre-trained text-to-image diffusion model $\epsilon_\theta$ using the DreamBooth algorithm. Unlike the Anti-DreamBooth problem, the attacker applies a purification function $\mathcal{P}$ to each sample before fine-tuning, aiming to remove adversarial noise:

\begin{equation}
    \tilde{x}^{(i)} = \mathcal{P}(x^{(i)} + \delta^{(i)}), \quad \tilde{\mathcal{X}}_{\text{db}} = \{\tilde{x}^{(i)}\}_{i=1}^{N_{\text{db}}}
\end{equation}

The attacker then optimizes the DreamBooth model on the purified set:
\begin{equation}
    \theta^* = \arg \min_{\theta} \sum_{i=1}^{N_{\text{db}}} \mathcal{L}_{\text{db}}(\theta, \tilde{x}^{(i)})
\end{equation}
Where $\mathcal{L}_{\text{db}}$ is the loss function on DreamBooth. Our goal is to evaluate the robustness of adversarial perturbations $\Delta_{\text{db}} = \{\delta^{(i)}\}_{i=1}^{N_{\text{db}}}$ under such purification, by measuring the residual personalization ability of the fine-tuned model $\epsilon_{\theta^*}$ with respect to the original identity set $\mathcal{X}$. Formally, we assess the following formula:

\begin{equation}
    \Delta^*_{\text{db}} = \arg \min_{\Delta_{\text{db}}} \mathcal{A}(\epsilon_{\theta^*}, \mathcal{X})
\end{equation}

\begin{equation*}
    \text{s.t.} \quad \theta^* = \arg \min_{\theta} \sum_{i=1}^{N_{\text{db}}} \mathcal{L}_{\text{db}}(\theta, \mathcal{P}(x^{(i)} + \delta^{(i)})), \quad \|\delta^{(i)}\|_p \leq \eta
\end{equation*}

Here, $\mathcal{A}(\cdot, \cdot)$ is some personalization evaluation function that measures how well the generated images preserve the target identity.

The Anti-DreamBooth framework introduces several settings that impose varying levels of constraints on the defenses. In its ``white-box" setting, the defender has full access to the pretrained text-to-image generator and the training prompt, resulting in the most powerful adversarial perturbations. Other settings weaken the strength of adversarial perturbations by introducing limitations, such as the defender being unaware of the training prompt used by the adversary or employing a different surrogate pretrained text-to-image generator. In our problem, we primarily focus on the white-box setting in the sense that if purification techniques can effectively filter out adversarial noise under this setting, they can also work in other settings where the adversarial perturbations are much weaker.

\subsection{Anti-Personalization Defenses}
\subsubsection{Adversarial Attacks.}
Adversarial attacks were originally introduced in classification tasks, where the goal is to add carefully crafted perturbations to the input such that a trained classifier produces incorrect predictions. When applied to generative models—such as diffusion models, the objective is to make the model believe the generated images as \textit{out-of-distribution (OOD)}.

To achieve this, existing methods adopt Projected Gradient Descent (PGD)~\cite{madry2018towards} to produce adversarial examples, \ie, \begin{equation}
    x^0 = x + \sigma,\; \mbox{where}\; \sigma\sim\mathcal{N}(0,1),\label{eq} 
\end{equation}
\begin{equation}
    x^{t+1} = \Pi_{x+\mathcal{S}}(x^t + \alpha sign(\nabla_x\mathcal{L}(\theta, x^t, y)),
\label{pgd}
\end{equation}    
where $x$ denotes the natural example and $x^0$ is obtained by perturbing $x$ with random noise $\sigma$ sampled from the normal distribution $\mathcal{N}(0,1)$, $t$ denotes the current time step, $\alpha$ is the step size, $\Pi$ denotes the projection function, $\mathcal{S} \subseteq \mathbb{R}^{d}$ denotes the  perturbation set of adversarial examples.

\subsubsection{Training Procedure.}

Existing Anti-DreamBooth defense mechanisms largely rely on the ASPL Alternating Surrogate and Perturbation Learning (ASPL) framework, which jointly optimizes the surrogate DreamBooth model and the adversarial perturbations to simulate real-world attackers.

In each ASPL iteration, the following three steps are performed:

\begin{enumerate}
    \item Clone and fine-tune the surrogate model: Create a copy of the current surrogate model $\epsilon_\theta$, denoted as $\epsilon'_{\theta'}$, and fine-tune it on the clean reference dataset $\mathcal{X}_A$ using the DreamBooth objective $\mathcal{L}_{\text{db}}$:
    \begin{equation}
        \theta' \leftarrow \arg\min_{\theta'} \sum_{x \in \mathcal{X}_A} \mathcal{L}_{\text{db}}(\theta', x)
    \end{equation}

    \item Update the adversarial perturbations: For each training sample $x^{(i)}$, optimize the perturbation $\delta^{(i)}$ to maximize the conditional generation loss $\mathcal{L}_{\text{cond}}$ with respect to the updated clone model:
    \begin{equation}
        \delta^{(i)} \leftarrow \arg\max_{\delta^{(i)}} \mathcal{L}_{\text{cond}}(\theta', x^{(i)} + \delta^{(i)})
    \end{equation}

    \item Update the original surrogate model: Using the updated adversarial examples, update the original surrogate model $\epsilon_\theta$ by minimizing the DreamBooth loss:
    \begin{equation}
        \theta \leftarrow \arg\min_{\theta} \sum_{i=1}^{N_{\text{db}}} \mathcal{L}_{\text{db}}(\theta, x^{(i)} + \delta^{(i)})
    \end{equation}
\end{enumerate}

This alternating optimization scheme enables the surrogate model to gradually adapt to the perturbed data distribution, thereby providing stronger guidance for learning transferable and robust protective perturbations. Improvements on Anti-DreamBooth focus on selecting specific timesteps within the denoising sequence for adversarial attacks~\cite{wang2024simac}, as well as designing a specialized loss function to disrupt the model’s guidance towards the target identity, such as employing the Cross-Attention Erasure mechanism~\cite{liu2024disrupting}. However, the basic idea still relies on PGD-based adversarial attacks and the ASPL framework.

\subsection{Purification Methods}
We categorize purification methods into two types: traditional filtering and adversarial purification.

\subsubsection{Traditional Filtering.}
We adopt a cascaded filtering pipeline consisting of repeated \textit{bilateral filtering} followed by \textit{guided filtering}, designed to remove adversarial patterns while preserving important structural information such as edges and facial features.

Specifically, given an input image $x \in \mathbb{R}^{H \times W \times 3}$, we first apply the bilateral filter iteratively:
\begin{equation}
    x^{(t+1)} = \text{BF}(x^{(t)}), \quad t = 0, \dots, T-1
\end{equation}
where $\text{BF}(\cdot)$ denotes the bilateral filter, which smooths the image while preserving edges based on both spatial proximity and pixel intensity similarity.

We then apply guided filtering to refine the result, using the original image $x$ as the guidance:
\begin{equation}
    x_{purified} = \text{GF}(x, x^{(T)})
\end{equation}
where $\text{GF}(\cdot)$ denotes the guided filter. The guided filter enhances structural consistency with the original content. Each of the two filters is applied for several iterations to enhance structural restoration, and they can be easily implemented with OpenCV, using \texttt{cv2.bilateralFilter} and \texttt{cv2.ximgproc.guidedFilter}.

\subsubsection{Adversarial Purification.} 
Adversarial purification aims to remove adversarial perturbations by leveraging generative priors from pretrained diffusion models. In our study, we use two representative methods: DiffPure and GrIDPure.

DiffPure is built upon stochastic differential editing (SDEdit)~\cite{mengsdedit}. Given a potentially adversarial image $x'$, DiffPure injects Gaussian noise into the image to obtain a noisy version $x_T$, and then denoises it through a pretrained diffusion model to reconstruct a purified image:

\begin{equation}
    x_T = \sqrt{\alpha_T} x' + \sqrt{1 - \alpha_T} \boldsymbol{\epsilon}, \quad \boldsymbol{\epsilon} \sim \mathcal{N}(0, \mathbf{I}),
\end{equation}
\begin{equation}
    x_{t-1} = \mathcal{D}_\theta(x_t, t), \quad \text{for } t = T, \dots, 1,
\end{equation}
where $\mathcal{D}_\theta$ denotes the denoising function parameterized by the diffusion model, and $\alpha_T$ denotes the noise level at timestep $T$ in the forward diffusion process, which controls the proportion of the original image retained versus the amount of noise injected. By gradually denoising from the noised state $x_T$, the method aims to push the adversarial image back to the clean data manifold.

When using large timesteps, DiffPure may distort the semantic content and structural details of the original image. GrIDPure addresses this limitation by introducing a high-resolution, structure-preserving purification framework built upon the core idea of DiffPure, with two key enhancements:

\begin{enumerate}
    \item Grid-wise purification: The image is partitioned into overlapping patches (grids), each of size $256 \times 256$, aligning with the input size of the diffusion model. This strategy enables the purification of high-resolution images (e.g., $512 \times 512$) while mitigating the loss of global contextual information.
    
    \item Small-step iterative denoising: Instead of using a large number of reverse diffusion steps once, GrIDPure performs multiple iterations of shallow denoising over small timestep intervals, which preserves fine-grained local structures and textures.
\end{enumerate}

Specifically, at each iteration $i$, the purified patch $\tilde{x}_i$ is blended with the previous output $x_i$ as follows:

\begin{equation}
    x_{i+1} = (1 - \gamma) \cdot \tilde{x}_i + \gamma \cdot x_i,
\end{equation}
where $\gamma$ is a blending factor that controls the trade-off between purification strength and image fidelity, after each patch is individually purified and overlapping regions are averaged, the entire image is iteratively refined.

\begin{figure}[]
\centering
\includegraphics[width=0.8\textwidth]{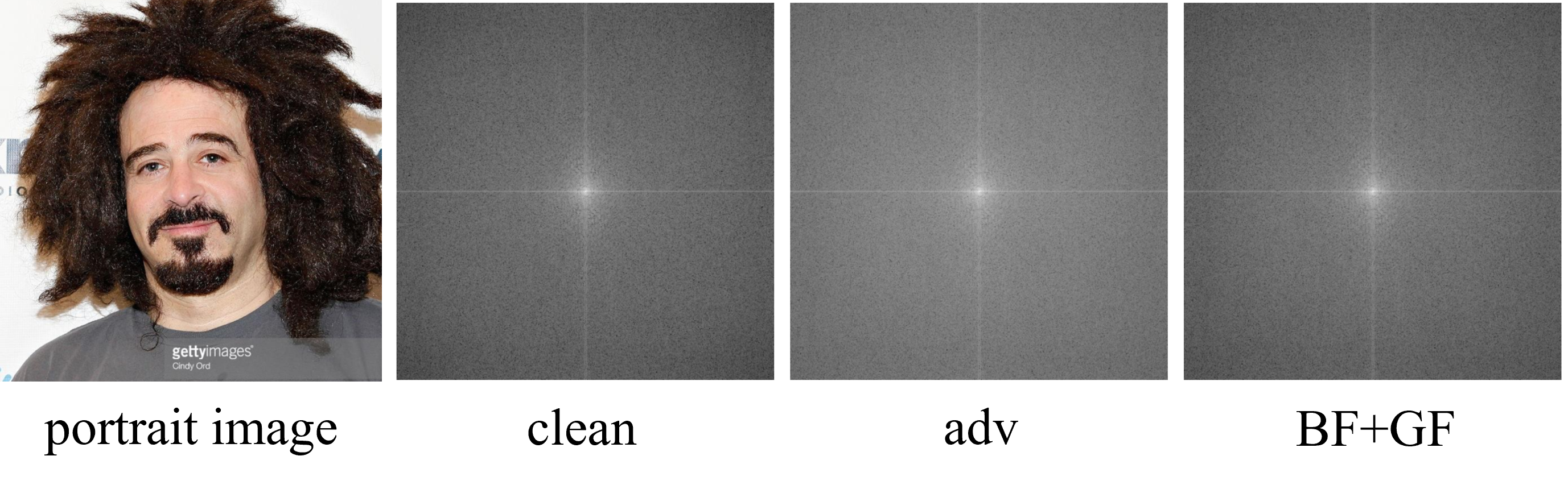} 
\caption{Fourier magnitude spectrum of clean images, adversarial examples generated by Anti-Dreambooth, and the corresponding purified images using bilateral filtering followed by guided filtering.}
\label{spectrum}
\end{figure}

\begin{figure}[]
\centering
\includegraphics[width=0.82\textwidth]{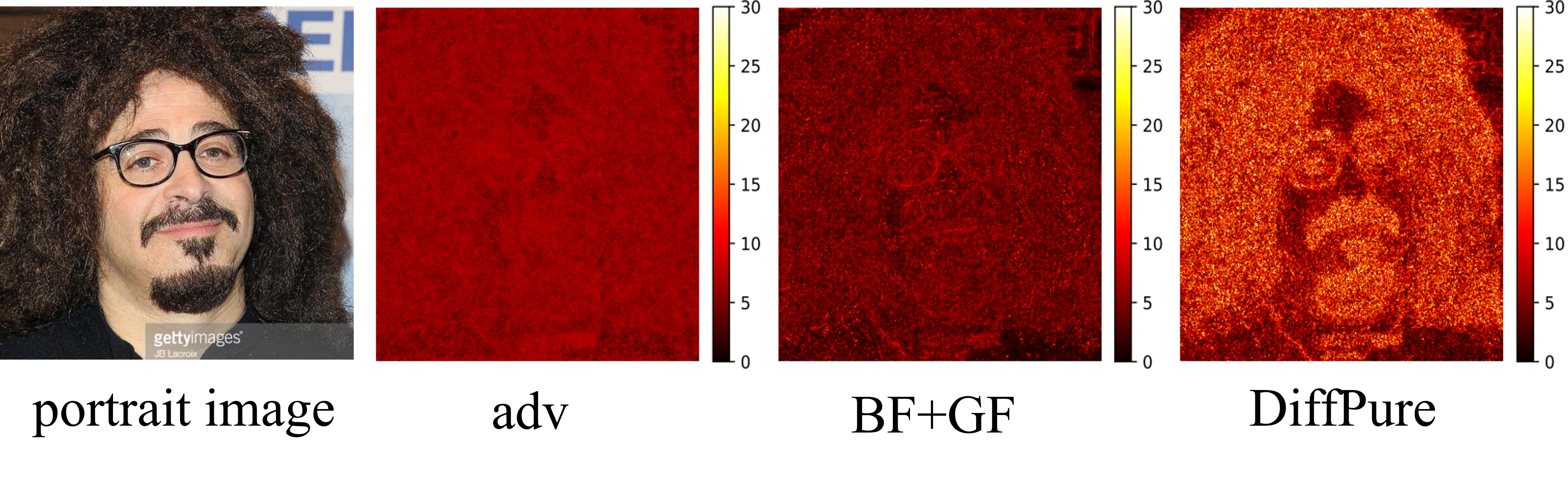} 
\caption{Visualization of clean images, heatmap of adversarial examples generated by Anti-Dreambooth, and their purified counterparts using bilateral filtering followed by guided filtering, and DiffPure.}
\label{heatmap}
\end{figure}

\subsection{Existing Anti-Personalization Adversarial Examples Fail Against Purification and Detection}
Although anti-personalization methods are effective at disrupting DreamBooth’s identity modeling, the adversarial examples they generate often exhibit limited robustness in practical scenarios. This fragility stems not only from the frequency-domain concentration of perturbations but also from the way these perturbations interact with image semantics. Specifically, to effectively disrupt identity learning, the attacks introduce structured distortions that alter global features and attention pathways in the model, making them visually noticeable, thus violating the principle of imperceptibility, and are susceptible to traditional filtering and adversarial purification. Especially, the PGD-based attacks employed by existing anti-personalizations introduce localized pixel-level perturbations that lack semantic alignment with the underlying image structure. As a result, traditional low-pass filters, which prioritize structural coherence and suppress abrupt intensity changes, can effectively filter out such noise patterns.

For adversarial purifications that aim to project adversarial inputs back onto the natural image manifold learned by pre-trained diffusion models. These methods have two advantages in purifying the adversarial examples by existing anti-personalization defenses:

\begin{itemize}
    \item Perturbation insensitivity: The initial forward diffusion step transforms the input into nearly pure noise, effectively erasing adversarial perturbations.
    \item Prior-constrained reconstruction: During the reverse denoising, the generative model reconstructs images aligned with its learned data distribution, thereby restoring semantic consistency despite the presence of adversarial noise.
\end{itemize}

We further highlight the perceptibility of adversarial perturbations generated by Anti-DreamBooth and the effectiveness of purification by visualizing the Fourier magnitude spectrum and heatmaps of corresponding images.
Fig.~\ref{spectrum} reveals that adversarial examples exhibit significantly stronger signal intensities compared to clean images and those purified using bilateral filtering followed by guided filtering. Fig.~\ref{heatmap} further demonstrates that adversarial examples largely lose the original image features, whereas both traditional filtering and adversarial purification methods can partially restore facial structures. These results indicate that adversarial perturbations by Anti-DreamBooth disrupt global structures, thereby increasing their perceptibility and detectability by both human observers and purification methods.

\begin{figure}[]
\centering
\includegraphics[width=0.72\textwidth]{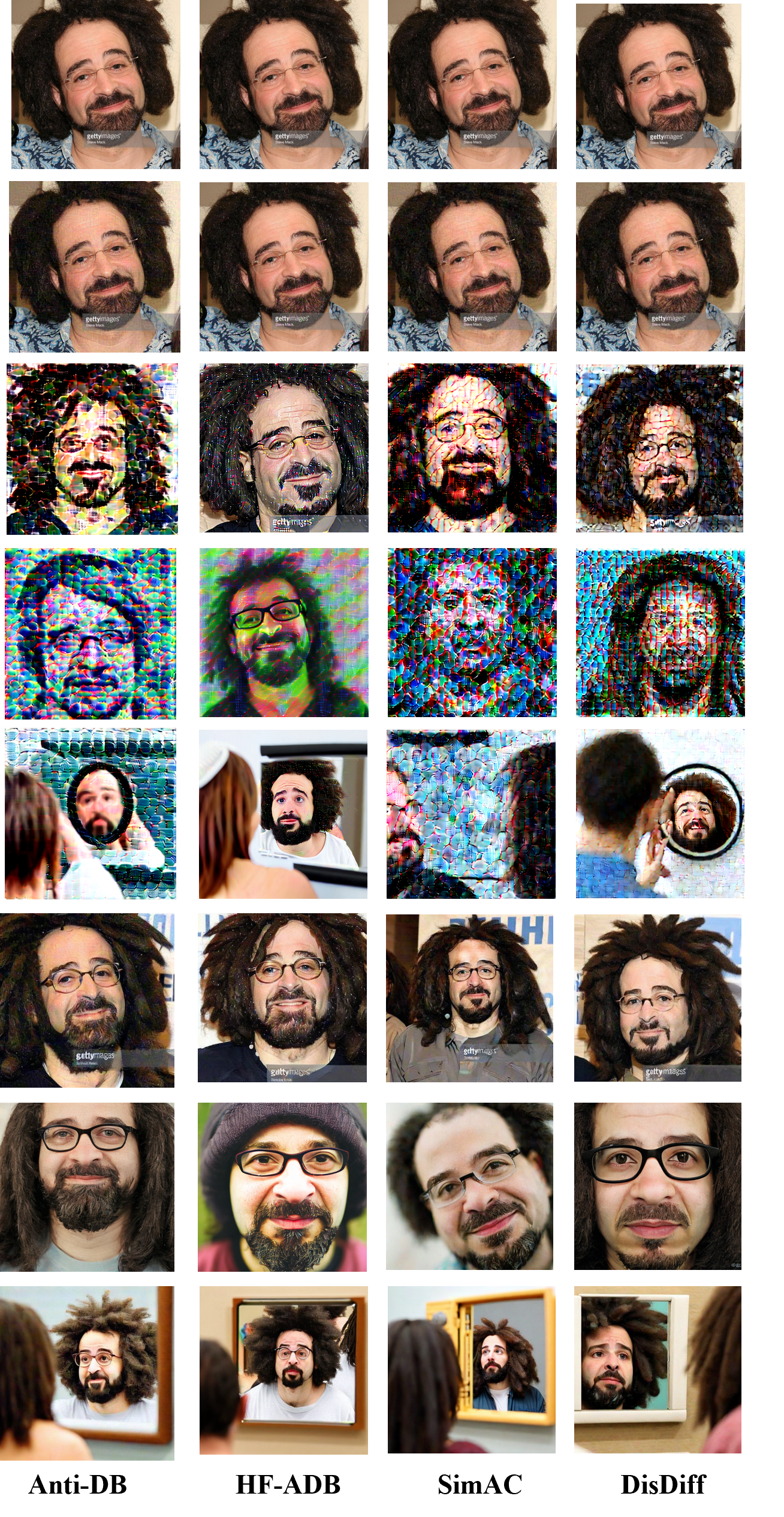} 
\caption{Visualization of clean portrait images (first row), adversarial examples generated by each anti-personalized method (second row), DreamBooth output on adversarial examples (middle three rows), and DreamBooth output on purified images using DiffPure (last three rows), text prompts are ``a photo of sks person", ``a dslr portrait of sks person", and ``a photo of sks person looking at the mirror".}
\label{visualize}
\end{figure}

\section{Experiments}

\subsection{Experimental Setup}
\subsubsection{Datasets.}
We evaluate on two widely adopted face datasets, CelebA-HQ~\cite{liu2015deep} and VGGFace2~\cite{cao2018vggface2}. We select some identities as the target subject for protection. All images are center-cropped and resized to 512×512 resolution, and the dataset is divided into set A, set B, and set C, and each set
contains 5 portrait images.

\subsubsection{Methods.} 
We evaluate the effectiveness of existing anti-personalization methods in protecting target subjects under purification-based post-processing. The protection methods include Anti-DreamBooth, HF-ADB, SimAC, and DisDiff. Specifically, we first generate adversarial examples using these approaches, and then apply three purification techniques: Bilateral Filtering + Guided Filtering, DiffPure, and GrIDPure to simulate realistic post-processing scenarios where an attacker attempts to remove adversarial noise.

\begin{table}[t]
\centering
\caption{Defense performance of existing anti-personalization methods before and after purifications}
\label{1}
\resizebox{1\textwidth}{!}{
\begin{tabular}{@{}c|c|cccc|cccc@{}}
\toprule
\multirow{2}{*}{\textbf{Method}} & \multirow{2}{*}{Purification} & \multicolumn{4}{c|}{“a photo of sks person”}                   & \multicolumn{4}{c}{“a dslr portrait of sks person”}            \\ \cmidrule(l){3-10} 
                                 &                               & FDFR↑         & ISM↓          & SER-FQA↓      & BRISQUE ↑      & FDFR↑         & ISM↓          & SER-FQA↓      & BRISQUE ↑      \\ \midrule
DreamBooth                       & -                             & 0.03          & 0.69          & 0.66          & 7.01           & 0.13          & 0.44          & 0.69          & 4.68           \\ \midrule
\multirow{4}{*}{Anti-DB}         & No                            & \textbf{0.27} & \textbf{0.40} & \textbf{0.1}  & 36.9           & \textbf{0.9}  & \textbf{0.19} & \textbf{0.25} & \textbf{36.6}  \\
                                 & BF+GF                         & 0             & 0.55          & 0.34          & 36.13          & 0.3           & 0.31          & 0.41          & 29.24          \\
                                 & DiffPure                      & 0             & 0.61          & 0.65          & 10.55          & 0.3           & 0.50          & 0.71          & -2.49          \\
                                 & GridPure                      & 0             & 0.66          & 0.65          & \textbf{46.17} & 0.2           & 0.41          & 0.63          & 17.24          \\ \midrule
\multirow{4}{*}{HF-ADB}          & No                            & 0             & \textbf{0.65} & 0.65          & 27.44          & 0.13          & 0.45          & \textbf{0.64} & \textbf{32.09} \\
                                 & BF+GF                         & 0             & 0.65          & 0.66          & 11.33          & 0.1           & 0.43          & 0.65          & 18.26          \\
                                 & DiffPure                      & 0             & 0.64          & \textbf{0.69} & 11.16          & 0.17          & 0.45          & 0.73          & 0.90           \\
                                 & GridPure                      & 0             & 0.63          & 0.61          & \textbf{56.96} & \textbf{0.26} & 0.44          & 0.65          & 21.01          \\ \midrule
\multirow{4}{*}{SimAC}           & No                            & \textbf{0.03} & \textbf{0.50} & \textbf{0.44} & \textbf{44.91} & \textbf{1}    & \textbf{-}    & \textbf{0.09} & \textbf{37.34} \\
                                 & BF+GF                         & 0             & 0.71          & 0.68          & 29.41          & 0.13          & 0.45          & 0.62          & 15.32          \\
                                 & DiffPure                      & 0             & 0.66          & 0.67          & 8.19           & 0.23          & 0.48          & 0.79          & 0.38           \\
                                 & GridPure                      & 0             & 0.67          & 0.66          & 37.85          & 0.3           & 0.44          & 0.68          & 20.58          \\ \midrule
\multirow{4}{*}{DisDiff}         & No                            & \textbf{0.07} & \textbf{0.57} & \textbf{0.27} & \textbf{36.26} & \textbf{0.87} & \textbf{0.21} & \textbf{0.22} & \textbf{37.06} \\
                                 & BF+GF                         & 0.03          & 0.70          & 0.69          & 7.81           & 0.17          & 0.47          & 0.66          & 17.23          \\
                                 & DiffPure                      & 0             & 0.67          & 0.68          & 5.62           & 0.23          & 0.45          & 0.76          & -1.51          \\
                                 & GridPure                      & 0             & 0.67          & 0.69          & 34.93          & 0.13          & 0.48          & 0.69          & 19.93          \\ \bottomrule
\end{tabular}
}
\end{table}

\subsubsection{Evaluation metrics.}

We use the following metrics to comprehensively evaluate whether the generation by DreamBooth can protect target identities:
We first use the Retinaface face detector~\cite{deng2020retinaface} to check whether a face is present in the generated image. Based on this, we compute the Face Detection Failure Rate (FDFR) accordingly, where a higher FDFR implies more effective identity protection. Subsequently, if a face is detected, we use ArcFace~\cite{deng2019arcface} to encode both the generated images and the clean images of the protected identity, and compute the cosine similarity between the two embeddings to assess the similarity of facial identity. This metric is Identity Score Matching (ISM), where a lower ISM indicates lower identity similarity, and thus stronger privacy protection. We also adopt two metrics to evaluate the quality of the generated images: SER-FIQ is used to assess the quality of the detected facial regions, while BRISQUE evaluates the overall image quality. Lower performance on these two metrics indicates stronger protection, as they reflect greater degradation in visual fidelity. 

To provide a more intuitive comparison of different methods on the same identity, we report the evaluation metrics for the subject n000050 in the VGGFace2 dataset. We also provide intuitive visual comparisons, and our findings consistently hold across other identities in the datasets.

\subsubsection{Implementation Details.}

For anti-personalization methods, we follow the default configurations specified by each approach. Specifically, we use Stable Diffusion v2.1 as the pre-trained generative backbone for DreamBooth fine-tuning. Both the text encoder and the UNet are fine-tuned with a batch size of 2, a learning rate of $5 \times 10^{-7}$, and a total of 1000 training steps. The instance prompt used during training is \textit{``a photo of sks person"}.

For the PGD attack, we set the step size \textbf{$\alpha = 0.005$} and the noise budget \textbf{$\eta = 0.05$}. All anti-personalization variants are optimized using the ASPL training strategy for 50 iterations. During inference, we use two text prompts: \textit{``a photo of sks person"} and \textit{``a dslr portrait of sks person"} for metric evaluation, and generate 30 images per prompt for quantitative metrics evaluation. For visualization, we use an additional text prompt \textit{``a photo of sks person looking at the mirror"}.

To evaluate the effectiveness of purifications on trained adversarial perturbations, we apply both traditional filters, \ie, bilateral filter followed by guided filter, and adversarial purification to remove the adversarial perturbations. The purified images are then used to fine-tune a new DreamBooth model. All experiments are implemented on a server with an NVIDIA A100 GPU (40GB).

\subsection{Quantitative Analysis}
To evaluate the effectiveness of existing anti-personalization methods before and after purification, we conduct a quantitative comparison using the two prompts listed in Table 1. For each prompt, we randomly sample 30 generated images and compute each metric, reporting the average results.

We highlight the best-performing metric for each method in Table~\ref {1}. For Anti-DreamBooth, SimAC, and DisDiff, the results are consistent: all purification techniques significantly degrade their protection across all evaluation metrics. Among the purification methods, DiffPure achieves the best overall performance, with performance that is closest to fine-tuning DreamBooth on clean images. Notably, even simple traditional filters exhibit strong purification capabilities, but fall short only in BRISQUE compared to DiffPure. In contrast, GrIDPure performs poorly on BRISQUE, possibly due to overpurification, which compromises image generation quality. Regarding HF-ADB, we observe that its protection is already poor even before purification. For instance, in terms of FDSR, most images generated by HF-ADB are successfully detected as containing faces, indicating weak identity protection.

We also provide visual comparisons in Fig.~\ref{visualize} to present the adversarial examples generated by different defense methods, as well as the outputs from DreamBooth fine-tuned on these adversarial examples and their purified counterparts via DiffPure. We find that adversarial examples produced by Anti-DreamBooth, SimAC, and DisDiff often exhibit noticeable semi-transparent artifacts, with SimAC being the most obvious (may need to zoom in on this figure). In contrast, HF-ADB introduces fine-grained ripple-like noise patterns. The DreamBooth outputs using purified images show that all methods produce facial images closely resembling the user's identity, indicating a failure of protection. However, we also notice that the adversarial noise from HF-ADB tends to persist even after purification, making it more difficult to remove compared to other approaches, but it does not provide effective protection even without purification.

\subsection{More Evaluations on SimAC with Another Identity}
Table~\ref{1} shows a representative identity commonly used in prior work. We further evaluate another identity (the one shown in Fig.~\ref{intro}) to illustrate that purification invalidates defense in the section, showing the same trend across different identities. Averaging results across identities is not a good choice since metric values vary by identity and prompt, yet the trend (purification degrades defenses) holds for different identities.

\subsubsection{Addtional Prompts.}
We further evaluate SimAC with two additional prompts: “a photo of sks person looking at the mirror” and “a photo of sks person in front of the Eiffel Tower”. From Table~\ref{add_prompts} and Fig.~\ref{additional_twp_prompts} (middle two rows), both visual and quantitative results consistently show that purification destroys SimAC across all prompts.

\begin{table}[]
\centering
\caption{Additional two prompts on SimAC before and after Diffpure}
\label{add_prompts}
\resizebox{0.66\textwidth}{!}{
\begin{tabular}{@{}c|cccc@{}}
\toprule
\multirow{2}{*}{Purification} & FDFR↑           & ISM↓           & SER-FQA↓          & BRISQUE ↑         \\ \cmidrule(l){2-5} 
                              & \multicolumn{4}{c}{“a photo of sks person looking at the mirror”}        \\ \midrule
No                            & 1               & –              & 0.004             & 39.65             \\
DiffPure                      & 0               & 0.40           & 0.56              & 6.18              \\ \midrule
\multicolumn{1}{l|}{}         & \multicolumn{4}{c}{"a photo of sks person in front of the Eiffel Tower"} \\ \midrule
No                            & 1               & -              & 0.07              & 40.86             \\
DiffPure                      & 0.17            & 0.19           & 0.52              & 9.94              \\ \bottomrule
\end{tabular}
}
\end{table}

\subsubsection{Combining SimAC and HF-ADB.}
\begin{figure}[!h]
\centering
\includegraphics[width=0.7\textwidth]{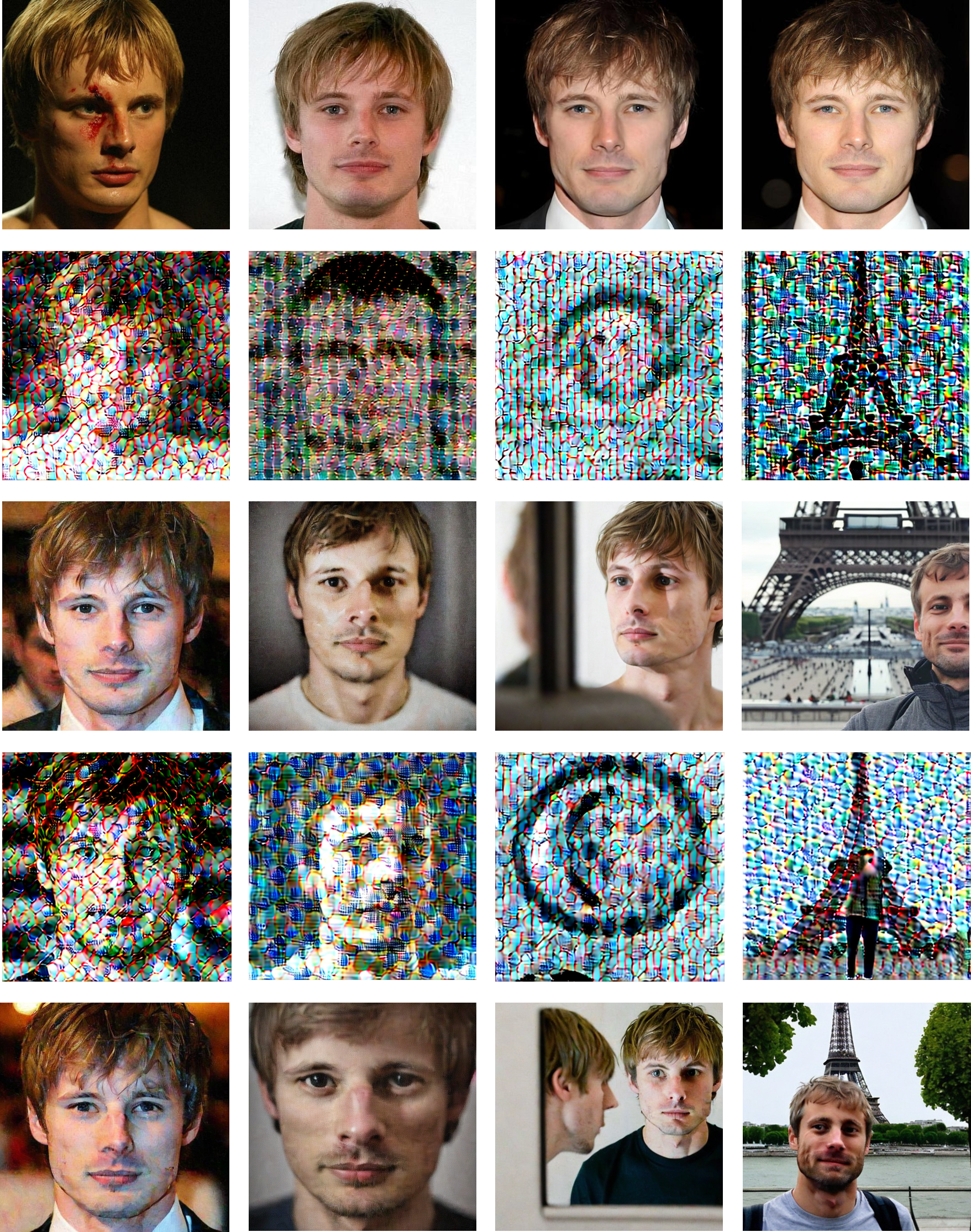} 
\caption{Visualization of clean portrait images (first row); DreamBooth output on SimAC generated adversarial examples and on their purified images using DiffPure(middle two rows); DreamBooth output on HF-SimAC generated adversarial examples and on their purified images using DiffPure(last two rows). Text prompts are ``a photo of sks person", ``a dslr portrait of sks person", ``a photo of sks person looking at the mirror", and ``a photo of sks person in front of the Eiffel Tower".}
\label{additional_twp_prompts}
\end{figure}

As the noise introduced by HF-ADB is more difficult to eliminate, we further combine it with SimAC, \ie, using SimAC while applying HF perturbations on the face mask regions. From Fig.~\ref{additional_twp_prompts} and Table~\ref{HF-SimAC}, before purification, HF slightly weakens the effect of SimAC in terms of visual perception, and the FDFR results also show that about 10\% more generated samples of HF-SimAC are recognized as containing a human compared to SimAC. Overall, however, the performance gap is marginal, indicating that both exhibit strong defense performance if no purification is applied; However, after purification, both defenses completely fail, and the fine-grained ripple-like noise in HF-ADB is also not observed in HF-SimAC. A plausible explanation is that, although HF perturbation is applied to the masked regions, the overall perturbation strength of SimAC dominates, thereby suppressing the effect of the HF perturbation.

\begin{table}[!t]
\centering
\caption{Quantitative results between SimAC and HF-SimAC before and after Diffpure}
\label{HF-SimAC}
\resizebox{0.64\textwidth}{!}{
\begin{tabular}{@{}c|c|cccc@{}}
\toprule
\multirow{2}{*}{Methods}  & \multirow{2}{*}{Purification} & FDFR↑   & ISM↓   & SER-FQA↓   & BRISQUE ↑   \\ \cmidrule(l){3-6} 
                          &                               & \multicolumn{4}{c}{“a photo of sks person”} \\ \midrule
\multirow{2}{*}{SimAC}    & No                            & 0.97    & 0.48   & 0.02       & 35.67       \\ \cmidrule(lr){2-2}
                          & DiffPure                      & 0       & 0.73   & 0.79       & 18.85       \\ \midrule
\multirow{2}{*}{HF-SimAC} & No                            & 0.87    & 0.41   & 0.08       & 38.57       \\ \cmidrule(lr){2-2}
                          & DiffPure                      & 0       & 0.66   & 0.76       & 8.37        \\ \bottomrule
\end{tabular}
}
\end{table}

\subsubsection{Purification Steps.}
\begin{figure}[h]
\centering
\includegraphics[width=0.7\textwidth]{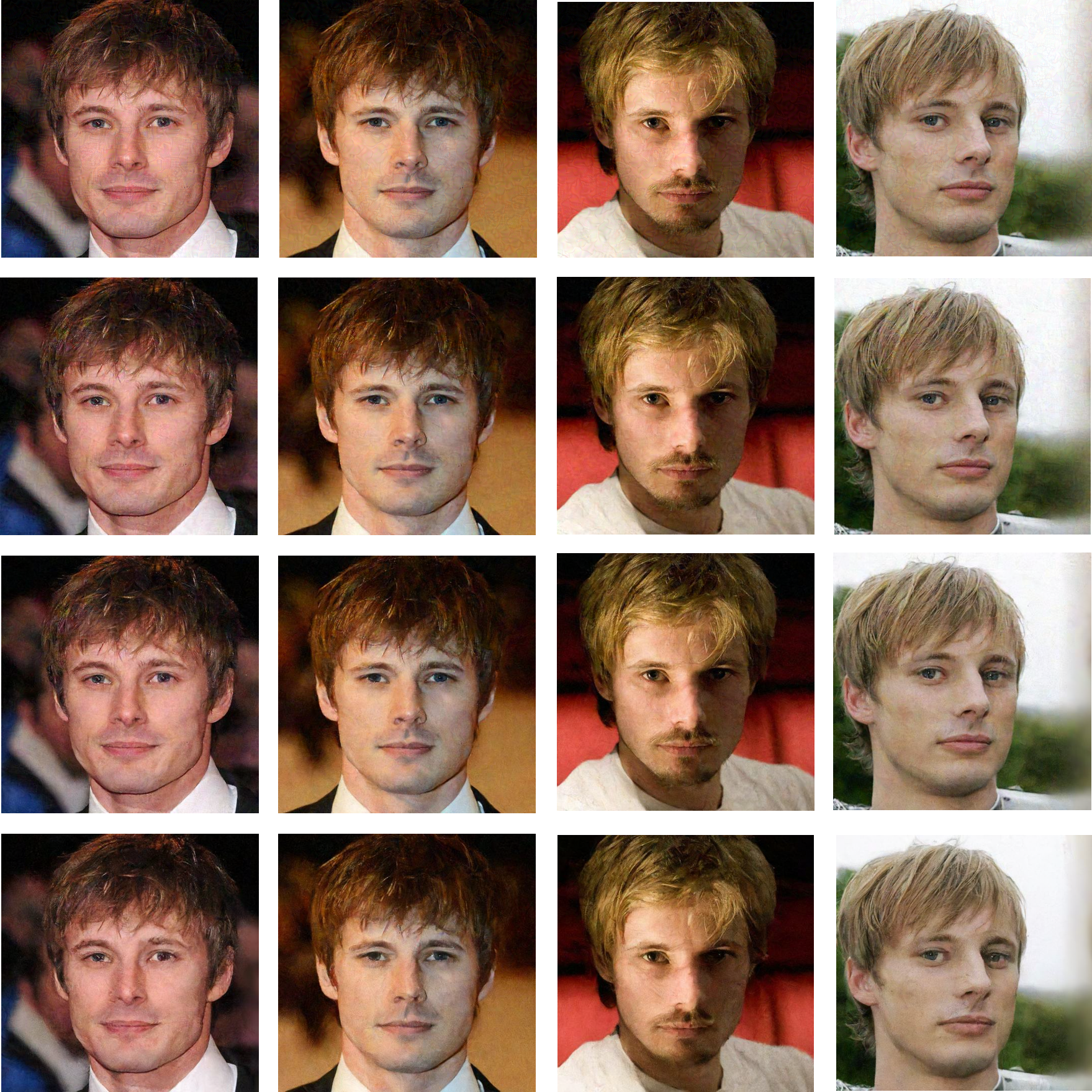} 
\caption{SimAC generated adversarial examples (first row); Purified examples by Diffpure with $t=20$; Purified examples by Diffpure with $t=50$; Purified examples by Diffpure with $t=100$ (Zoom in for better comparison).}
\label{diffpure_compare}
\end{figure}

In DiffPure, the purification step $t$ controls the purification strength. We primarily focus on a realistic purification threat, \ie, the default parameter ($t=50$) in our main experiments, which is sufficient to expose the fragile pattern with very few computational resources. Additionally, we evaluate $t=20$ and $t=100$ in this section (Table~\ref{diffpure_compare}) to show that purification remains effective for typical parameter ranges, without requiring complex tuning. The average processing time per image is 9.45s ($t=20$), 4.33s ($t=50$), and 8.10s ($t=100$). GPU memory usage is about 4 GB for all settings. 

\section{Conclusion}
In this paper, we introduce a more realistic and challenging evaluation paradigm for anti-personalization defenses: whether they can still effectively protect a user’s facial identity after purification. Unfortunately, our findings reveal a consistent failure across all existing methods. Once the adversarial examples are processed through either traditional filtering techniques or adversarial purification methods, the target user’s identity can be reconstructed by DreamBooth, completely bypassing the intended protection.

We also observe that the high-frequency perturbations introduced by HF-ADB tend to be more resilient to purification, suggesting a potential direction for future research. Although HF-ADB falls short in providing effective identity protection, its underlying strategy of spatially and spectrally varying perturbations, \ie, injecting different perturbations into different image regions, may serve as a solution for this challenge. Additionally, a deeper theoretical analysis and understanding of how purification interacts with identity-preserving features may also enable more reliable protection mechanisms. We hope our findings will encourage future work toward building stronger protection solutions for personalized image generation.

\section{Acknowledgements}

XZ's contribution is supported by the UK EPSRC New Investigator Award [EP/Z536568/1] and NVIDIA Academic Grant Program. The corresponding author is WR (w.ruan@trustai.uk). ZC, YZ, and XY's contributions are supported by the China Scholarship Council.

\section{Ethics Statement}

In this paper, we aim to reveal that existing anti-personalization methods are vulnerable to purification and encourage stronger, privacy-preserving defenses for personalized generative models, rather than encouraging the misuse of purification techniques.

\bibliographystyle{splncs04}
\bibliography{mybibliography}

\end{document}